\documentclass[twocolumn,showpacs,preprintnumbers,amsmath,amssymb]{revtex4}

\usepackage{graphicx}
\usepackage{dcolumn}
\usepackage{bm}

\begin{document}

\preprint{}

\title{Efficient Decision-Making by Volume-Conserving Physical Object}

\author{Song-Ju Kim$^{1}$}
\email{KIM.Songju@nims.go.jp}

\author{Masashi Aono$^{2}$}

\author{Etsushi Nameda$^{3}$}

\affiliation{%
$^{1}$WPI Center for MANA, National Institute for Materials Science, Tsukuba, Ibaraki 305--0044, Japan
}%

\affiliation{%
$^{2}$Earth-Life Science Institute, Tokyo Institute of Technology, Tokyo 152--8550 \& PRESTO JST, Japan
}%

\affiliation{%
$^{3}$RIKEN, 2--1 Hirosawa, Wako, Saitama 351--0198, Japan
}%

\date{\today}

\begin{abstract}
We demonstrate that any physical object, as long as its volume is conserved when coupled with suitable operations, provides a sophisticated decision-making capability.
We consider the problem of finding, as accurately and quickly as possible, the most profitable option from a set of options that gives stochastic rewards.
These decisions are made as dictated by a physical object, which is moved in a manner similar to the fluctuations of a rigid body in a tug-of-war game. 
Our analytical calculations validate statistical reasons why our method exhibits higher efficiency than conventional algorithms.
\end{abstract}

\pacs{45.40.-f, 89.20.Ff, 89.20.Kk}
                             
\maketitle

The computing principles in modern digital paradigms have been designed to be dissociated from the underlying physics of natural phenomena~\cite{beyondturing}.
In the construction of CMOS devices, wide-band-gap materials have been employed so that physical fluctuations such as thermal noise, which often violate logically-valid behavior, could be neglected~\cite{cmos}.
Since electron dynamics constrained by physical laws cannot be controlled when only parameters of the same degree of freedom as those of logical input--output responses are modulated, considerably complicated circuits are required for implementing relatively simple logic gates such as NAND and NOR~\cite{cmos2}.
However, these efforts to circumvent the division between physics and computation are costly in terms of energy consumption and manufacturing resources.
On the other hand, when we look at the natural world, information processing in biological systems is elegantly coupled with their underlying physics~\cite{natcom,natcom2}.
This suggests a potential for establishing a new physics-based analog-computing paradigm.
In this Letter, we show that a physical constraint, the conservation law for the volume of a rigid body, allows for efficient solving of decision-making problems when subjected to suitable operations involving fluctuations.

Suppose there are $M$ slot machines, each of which returns a reward; for example, a coin, with a certain probability that is unknown to a player.
Let us consider a minimal case: two machines A and B give rewards with individual probabilities $P_A$ and $P_B$, respectively.
The player makes a decision on which machine to play at each trial, trying to maximize the total reward obtained after repeating several trials.
The multi-armed bandit problem (MBP) is used to determine the optimal strategy for finding the machine with the highest reward probability as accurately and quickly as possible by referring to past experiences.

The MBP is formulated as a mathematical problem without loss of generality and so is related to various stochastic phenomena.
In fact, many application problems in diverse fields, such as communications (cognitive networks~\cite{cog,cog2}), commerce (advertising on the web~\cite{web}), entertainment (Monte-Carlo tree search, which is used for computer games~\cite{uct,mogo}), and so on, can be reduced to MBPs.
Particularly, the ``upper confidence bound 1 (UCB1) algorithm'' for solving MBPs is used worldwide in many practical applications~\cite{auer}.

In the context of reinforcement learning, the MBP was originally described by Robbins~\cite{robbins}, though the essence of the problem had been studied earlier by Thompson~\cite{thompson}.
The optimal strategy, called the ``Gittins index'', is known only for a limited class of problems in which the reward distributions are assumed to be known to the players~\cite{gittins1,gittins2}.
Even in this limited class, in practice, computing the Gittins index becomes intractable for many cases.  
For the algorithms proposed by Agrawal and Auer et al., another index was expressed as a simple function of the reward sums obtained from the machines~\cite{agra,auer}.

Kim et al. proposed an MBP solution using a dynamical system, called ``tug-of-war (TOW) dynamics''; this algorithm was inspired by the spatiotemporal dynamics of a single-celled amoeboid organism (the true slime mold {\it P. polycephalum})~\cite{kim1,kim2,kim3,kim4,kim5,kim6}, which maintains a constant intracellular-resource volume while collecting environmental information by concurrently expanding and shrinking its pseudopod-like terminal parts.
In this nature-inspired algorithm, the decision-making function is derived from its underlying physics, resembling that of a tug-of-war game.
The physical constraint in TOW dynamics, the conservation law for the volume of the amoeboid body, entails a nonlocal correlation among the terminal parts, that is, the volume increment in one part is immediately compensated by volume decrement(s) in the other part(s). 
In our previous studies~\cite{kim1,kim2,kim3,kim4,kim5,kim6}, we showed that, owing to the nonlocal correlation derived from the volume-conservation law, TOW dynamics exhibit higher performance than other well-known algorithms such as the modified $\epsilon$-greedy algorithm and the modified softmax algorithm, which is comparable to the UCB1-tuned algorithm (seen as the best choice among parameter-free algorithms~\cite{auer}). 
These observations suggest that efficient decision-making devices could be implemented using any physical object as long as it held some common physical attributes such as the conservation law.
In fact, Kim et al. demonstrated that optical energy-transfer dynamics between quantum dots, in which energy is conserved, can be exploited for the implementation of TOW dynamics~\cite{QDM,QDM2}.

\begin{figure}[h]
\centering
\includegraphics[height=50mm]{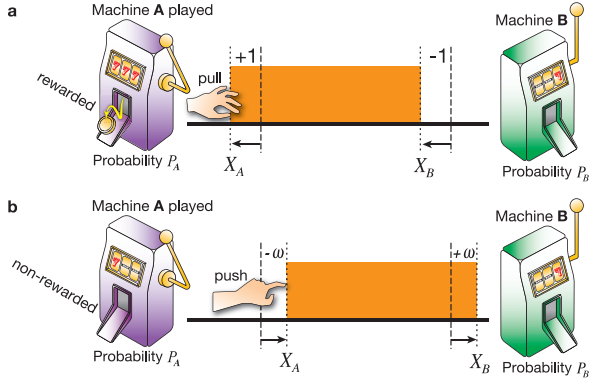}
\caption{TOW dynamics. If machine $k$ ($k\in \{A,B\}$) is played at each time $t$, $+1$ and $-\omega$ are added to $X_k(t-1)$ for rewarding~(a) and non-rewarding cases (b), respectively.}
\label{fig:stow}
\end{figure}

Consider a volume-conserving physical object; for example, a rigid body like an iron bar (the slot-machine's handle), as shown in Fig.~\ref{fig:stow}.
Here, the variable $X_k$ represents the displacement of terminal $k$ from an initial position, where $k\in \{A,B\}$.
If $X_k$ is a maximum, we assume that the body makes a decision to play machine $k$.
In TOW dynamics, the MBP is represented in its inverse form: instead of ``rewarding'' the player when machine $k$ produces a coin with a probability $P_k$, we ``punish'' the player when the machine gives no coin with a probability $1 - P_k$.
In this respect, the displacement $X_A$ ($= - X_B$) is determined by the following equations:
\begin{eqnarray}
X_A(t) & = & Q_A(t) - Q_B(t) + \delta(t) ,\label{diffe}\\
Q_k(t) & = & N_k - (1 + \omega)\hspace{1mm}L_k . \label{eq:org}
\end{eqnarray}
Here, $Q_k(t)$ ($k\in \{A,B\}$) is an ``estimate'' of information on past experiences accumulated from the initial time $1$ to current time $t$, $N_k$ counts the number of times that machine $k$ has been played, $L_k$ counts the number of punishments when playing machine $k$, $\delta(t)$ is an arbitrary fluctuation to which the body is subjected, and $\omega$ is a weighting parameter to be described in detail later on in this Letter.
Eq.(\ref{eq:org}), called the ``learning rule'', reflects the volume-conservation law. Consequently the TOW dynamics evolve according to a particularly simple rule: in addition to the fluctuation, if machine $k$ is played at each time $t$, $+1$ and $-\omega$ are added to $X_k(t-1)$ when rewarded and non-rewarded, respectively (Fig.~\ref{fig:stow}).


\begin{figure}[h]
\centering
\includegraphics[height=50mm]{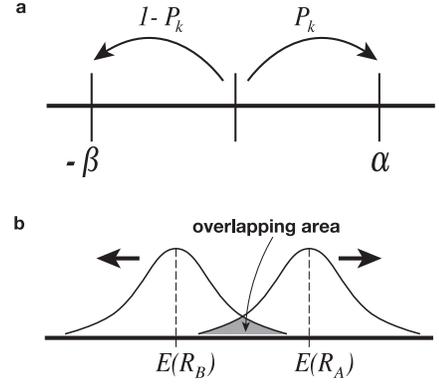}
\caption{(a) Random walk: flight $\alpha$ when rewarded with $P_k$ or flight $-\beta$ when non-rewarded with $1 - P_k$. (b) Probability distributions of two random walks.}
\label{fig:random}
\end{figure}

To explore the origins of the high performance of TOW dynamics, let us consider a random-walk model for comparison.
As shown in Fig.~\ref{fig:random}(a), $\alpha$ (right flight when rewarded) and $\beta$ (left flight when non-rewarded) are the parameters.
We assume that $P_A$ $>$ $P_B$ for simplicity. 
After time step $t$, the displacement $R_k(t)$ ($k\in \{A,B\}$) can be described by
\begin{eqnarray}
R_k(t) & = & \alpha (N_k - L_k) - \beta \hspace{1mm}L_k  \nonumber \\
       & = & \alpha N_k - (\alpha + \beta)\hspace{1mm}L_k . \label{eq:ran}
\end{eqnarray}
The expected value of $R_k$ can be obtained from the following equation:
\begin{equation}
E( R_k(t) ) = \{\alpha P_k - \beta (1 - P_k)\}\hspace{1mm} N_k .
\end{equation}

In the overlapping area between the two distributions shown in Fig.~\ref{fig:random}(b), we cannot accurately estimate which is larger.
The overlapping area should decrease as $N_k$ increases so as to avoid incorrect judgments.  
This requirement can be expressed by the following forms:
\begin{eqnarray}
  \alpha P_A - \beta (1 - P_A)     & > & 0 ,  \\
  \alpha P_B - \beta (1 - P_B)     & < & 0 . \label{eq:cond1}
\end{eqnarray}
These expressions can be rearranged into the form  
\begin{equation}
P_B  < \frac{\beta}{\alpha + \beta}   < P_A .  \label{eq:cond1}
\end{equation}
In other words, the parameters $\alpha$ and $\beta$ must satisfy the above conditions so that the random walk correctly represents the larger judgment.

We can easily confirm that the following form satisfies the above conditions:
\begin{equation}
\frac{\beta}{\alpha + \beta}   = \frac{P_A + P_B}{2} .\label{eq:fixed}
\end{equation}
From $R_k(t)/\alpha $ $=$ $Q_k(t)$, we obtain $\omega = \frac{\beta}{\alpha}$. 
From this and Eq.(\ref{eq:fixed}), we obtain
\begin{eqnarray}
\omega_0 & = & \frac{\gamma}{2 - \gamma} , \label{eq:w0}\\
\gamma & = & P_A + P_B .
\end{eqnarray}
Here, we have set the parameter $\omega$ to $\omega_0$.
Therefore, we can conclude that the algorithm using the learning rule $Q_k$ with the parameter $\omega_0$ can solve the MBP correctly.

In many popular algorithms such as the $\epsilon$-greedy algorithm, at each time $t$, an estimate of reward probability is updated for either of the two machines being played.
On the other hand, in an imaginary circumstance in which the sum of the reward probabilities $\gamma$ $=$ $P_A$ $+$ $P_B$ is known to the player, we can update both of the two estimates simultaneously, even though only one of the machines was played.

\begin{table}[h]
\caption{Estimates for each reward probability based on the knowledge that machine $A$ was played $N_A$ times and that machine $B$ was played $N_B$ times---on the assumption that the sum of the reward probabilities $\gamma$ $=$ $P_A$ $+$ $P_B$ is known.}
\label{table:1}
\begin{center}
\begin{tabular}{|c|c|c|c|}\hline \hline
$A$: & $\frac{N_A - L_A}{N_A}$ & $B$: & $\gamma \hspace{1mm} -$ $\frac{N_A - L_A}{N_A}$ \\ \hline
$A$: & $\gamma \hspace{1mm} -$ $\frac{N_B - L_B}{N_B}$ & $B$: & $\frac{N_B - L_B}{N_B}$ \\ \hline
\end{tabular}
\end{center}
\end{table}
The top and bottom rows of Table~\ref{table:1} provide estimates based on the knowledge that machine $A$ was played $N_A$ times and that machine $B$ was played $N_B$ times, respectively.
Note that we can also update the estimate of the machine that was not played, owing to the given $\gamma$.

From the above estimates, each expected reward $Q^{\prime}_k$ ($k\in \{A,B\}$) is given as follows:
\begin{eqnarray}
Q^{\prime}_A & = & N_A \hspace{1mm} \frac{N_A - L_A}{N_A} + N_B \hspace{1mm} \bigl( \gamma \hspace{1mm} - \frac{N_B - L_B}{N_B} \bigr) \nonumber \\
 & = & N_A - L_A + (\gamma - 1) \hspace{1mm} N_B + L_B, \label{eq:qAp}\\
Q^{\prime}_B & = & N_A\hspace{1mm} \bigl( \gamma \hspace{1mm} - \frac{N_A - L_A}{N_A} \bigr) + N_B\hspace{1mm} \frac{N_B - L_B}{N_B} \nonumber \\
& = & N_B - L_B + (\gamma - 1) \hspace{1mm} N_A + L_A. \label{eq:qBp}
\end{eqnarray}
These expected rewards, $Q^{\prime}_j$s, are not the same as those given by the learning rules of TOW dynamics, $Q_j$s in Eq.(\ref{eq:org}).  
However, what we use substantially in TOW dynamics is the difference
\begin{equation}
Q_A - Q_B  =  (N_A - N_B) - (1 + \omega)\hspace{1mm}(L_A - L_B)\label{eq:dq}.
\end{equation}
When we transform the expected rewards $Q^{\prime}_j$s into $Q^{\prime \prime}_j  =  Q^{\prime}_j / (2 - \gamma)$, 
we can obtain the difference
\begin{equation}
Q^{\prime \prime}_A - Q^{\prime \prime}_B  =  (N_A - N_B) - \frac{2}{2-\gamma} \hspace{1mm} (L_A - L_B). \label{eq:dqpp}
\end{equation}
Comparing the coefficients of Eq.(\ref{eq:dq}) and (\ref{eq:dqpp}), the differences in their constituent terms are always equal when $\omega=\omega_0$ (Eq.(\ref{eq:w0})) is satisfied.
Eventually, we can obtain the nearly optimal weighting parameter $\omega_0$ in terms of $\gamma$.

This derivation implies that the learning rule for TOW dynamics is equivalent to that of the imaginary system in which both of the two estimates can be updated simultaneously.
In other words, TOW dynamics imitates the imaginary system that determines its next move at time $t+1$ in referring to the estimates of the two machines, even if one of them was not actually played at time $t$. 
This unique feature in the learning rule, derived from the fact that the sum of reward probabilities is given in advance, may be one of the origins of the high performance of TOW dynamics.

Monte Carlo simulations were performed it was verified that TOW dynamics with $\omega_0$ exhibits an exceptionally high performance, which is comparable to its peak performance---achieved with the optimal parameter $\omega_{opt}$.
To derive the optimal value $\omega_{opt}$ accurately, we need to take into account the fluctuation and other dynamics of terminals~\cite{kim4}.

In addition, the essence of the process described here can be generalized to $M$-machine cases.
To separate distributions of the top $m$-th and top $(m+1)$-th machine, as shown in Fig.~\ref{fig:random}(b), all we need is the following $\omega_0$:
\begin{eqnarray}
\omega_0  & = & \frac{\gamma^{\prime}}{2 - \gamma^{\prime}} , \\
\gamma^{\prime} & = & P_{(m)} + P_{(m+1)}
\end{eqnarray} 
Here, $P_{(m)}$ denotes the top $m$-th reward probability.
In fact, for $M$-machine and $X$-player cases, we have designed a physical system that can  determine the overall optimal state, called the ``social maximum,'' quickly and accurately~\cite{bom}.


To further investigate the origins of the high performance of TOW dynamics, let us consider another imaginary model for solving the MBP, called the ``cheater algorithm.''
The cheater algorithm selects a machine to play according to the following estimate $S_k$ ($k\in \{A,B\}$)
\begin{eqnarray}
S_A  & = & X_{A, 1} + X_{A, 2}, + \cdots + X_{A, N} , \\
S_B  & = & X_{B, 1} + X_{B, 2}, + \cdots + X_{B, N} .
\end{eqnarray} 
Here, $X_{k, i}$ is a random variable that takes either $1$ (rewarded) or $0$ (non-rewarded).
If $S_A$ $>$ $S_B$ at time $t=N$, machine $A$ is played at time $t=N+1$. 
If $S_B$ $>$ $S_A$ at time $t=N$, machine $B$ is played at time $t=N+1$. 
If $S_A$ $=$ $S_B$ at time $t=N$, a machine is played randomly at time $t=N+1$. 
Note that the algorithm refers to results of both machines at time $t$ without any attention to which machine was played at time $t-1$.
In other words, the algorithm ``cheats'' because it plays both machines and collects both results, but declares that it plays only one machine at a time.   

The expected value and the variance of $X_k$ are defined as $E (X_k)  =  \mu_k $ and $V (X_k)  =  \sigma_k^{2}$.
Here, $\mu_k$ is the same as the $P_k$ defined earlier.
From the central-limit theorem, $S_k$ has a Gaussian distribution with $E (S_k)  =  \mu_k N$ and $V (S_k)  =  \sigma_k^{2} N$. 
If we define a new variable $S = S_A - S_B$, 
$S$ has a Gaussian distribution and carries the following values:
\begin{eqnarray}
E (S)  & = & (\mu_A + \mu_B) N  ,\\
V (S)  & = & (\sigma_A^{2} +  \sigma_B^{2}) N ,\\
\sigma (S)  & = & \sqrt{ \sigma_A^{2} +  \sigma_B^{2}} \sqrt{N} .
\end{eqnarray} 

\begin{figure}[h]
\centering
\includegraphics[height=30mm]{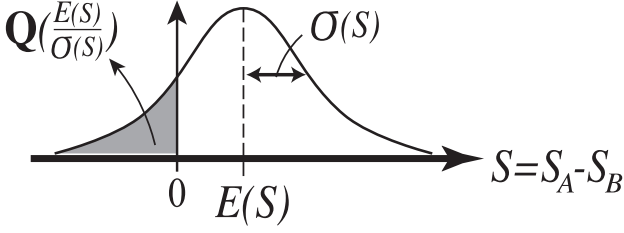}
\caption{{\bf Q}$(\frac{E(S)}{\sigma(S)})$: probability of selecting the lower-reward machine using the cheater algorithm}
\label{fig:Qfunc}
\end{figure}
 
From Fig.~\ref{fig:Qfunc}, the probability of playing machine $B$, which has a lower reward probability, can be described as {\bf Q}$(\frac{E(S)}{\sigma(S)})$.
Here, {\bf Q}$(x)$ is a {\bf Q}-function.
We obtain 
\begin{eqnarray}
P(t=N+1,B) & = & {\bf Q}(\phi \sqrt{N}) .
\end{eqnarray}
Here, $\phi = \frac{\mu_A - \mu_B}{\sqrt{\sigma_A^2 + \sigma_B^2}}$.

Using the Chernoff bound ${\bf Q}(x)  \leq  \frac{1}{2} \exp(-\frac{x^2}{2})$, 
we can calculate the upper bound of a measure, called the ``regret'', which quantifies the accumulated losses of the cheater algorithm.
\begin{equation}
regret = (\mu_A - \mu_B) E(N_B).
\end{equation}
\begin{eqnarray}
E(N_B) & = & \Sigma_{t=0}^{N-1} {\bf Q}(\phi \sqrt{t}) \nonumber \\
       & \leq & \Sigma_{t=0}^{N-1} \frac{1}{2} \exp(-\frac{\phi^2}{2} t) \nonumber \\
       & = & \frac{1}{2} + \Sigma_{t=1}^{N-1} \frac{1}{2} \exp(-\frac{\phi^2}{2} t) \nonumber \\
       & \leq & \frac{1}{2} + \int_{0}^{N-1} \frac{1}{2} \exp(-\frac{\phi^2}{2} t) dt \nonumber \\
       & = & \frac{1}{2} - \frac{1}{\phi^2} \left( \exp(-\frac{\phi^2}{2}(N-1))  - 1  \right) \\
       & \rightarrow & \frac{1}{2} + \frac{1}{\phi^2} .
\end{eqnarray} 
Note that the regret becomes constant as $N$ increases.

Using the ``cheated'' results, we can also calculate the regret of TOW dynamics in the same way.
In this case, 
\begin{eqnarray}
S_A  & = & X_{A, 1} + X_{A, 2}, + \cdots + X_{A, N_A} - \omega L_A , \\
S_B  & = & X_{B, 1} + X_{B, 2}, + \cdots + X_{B, N_B} - \omega L_B .
\end{eqnarray} 
$X_{k, i}$ is also a random variable that takes either $1$ (rewarded) or $0$ (non-rewarded).
Here, we use $L_k$$=$$(1 - \mu_k) N_k$.
Then, we obtain $E (S_k)  =  \{ \mu_k - (1 - \mu_k) \omega \}  N_k$ and $V (S_k)  = \sigma_k^{2} N_k$. 
Using the new variables $S = S_A - S_B$, $N = N_A + N_N$, and $D = N_A - N_N$, we also obtain 
\begin{eqnarray}
E (S)  & = & \frac{\mu_A - \mu_B}{2} (1 + \omega) N + \nonumber \\
       &   & \{ \frac{\mu_A + \mu_B}{2} (1 + \omega) - \omega \} D   ,\\
V (S)  & = & \frac{\sigma_A^2 + \sigma_B^2}{2} N +  \frac{\sigma_A^2 - \sigma_B^2}{2} D .
\end{eqnarray}

If the conditions $\omega  = \omega_0$ and $\sigma_A = \sigma_B$ $\equiv \sigma$ are satisfied, 
we then obtain 
\begin{eqnarray}
E (S)  & = & \frac{\mu_A - \mu_B}{2} (1 + \omega_0) N ,\\
V (S)  & = & {\sigma^2} N ,
\end{eqnarray} 
and 
\begin{eqnarray}
P(t=N+1,B) & = & {\bf Q}(\phi_{T} \sqrt{N}) .   
\end{eqnarray}
Here, $\phi_{T} = \frac{(\mu_A - \mu_B) (1 + \omega_0)}{2 \sigma}$.

We can then calculate the upper bound of the regret for TOW dynamics
\begin{eqnarray}
E(N_B) & = & \Sigma_{t=0}^{N-1} {\bf Q}(\phi_{T} \sqrt{t}) \nonumber \\
       & \leq & \frac{1}{2} - \frac{1}{\phi_{T}^2} \left( \exp(-\frac{\phi_{T}^2}{2}(N-1))  - 1  \right) \\
       & \rightarrow & \frac{1}{2} + \frac{1}{\phi_{T}^2} .
\end{eqnarray} 
Note that the regret for TOW dynamics also becomes constant as $N$ increases.

It is known that optimal algorithms for the MBP, defined by Auer et al., have a regret proportional to $\log(N)$. 
The regret has no finite upper bound as $N$ increases because it continues to require playing the lower-reward machine to ensure that the probability of incorrect judgment goes to zero.  
A constant regret means that the probability of incorrect judgment remains non-zero in TOW dynamics, although this probability is nearly equal to zero.
However, it would appear that the reward probabilities change frequently in actual decision-making situations, and their long-term behavior is not crucial for many practical purposes.
For this reason, TOW dynamics would be more suited to real-world applications.

In this Letter, we proposed TOW dynamics for solving the MBP and analytically validated that their high efficiency in making a series of decisions for maximizing the total sum of stochastically obtained rewards is embedded in any volume-conserving physical object when subjected to suitable operations involving fluctuations.
In conventional decision-making algorithms for solving the MBP, the parameter for adjusting the ``exploration time'' must be optimized.
This exploration parameter often reflects the difference between the rewarded experiences, i.e., $|P_A - P_B |$.
In contrast, TOW dynamics demonstrates that a higher performance can be achieved by introducing a weighting parameter $\omega_0$ that refers to the sum of the rewarded experiences, i.e., $P_A$ $+$ $P_B$.
Owing to this novelty, the high performance of TOW dynamics can be reproduced when implementing these dynamics with various volume-conserving physical objects.
Thus, our proposed physics-based analog-computing paradigm would be useful for a variety of real-world applications and for understanding the biological information-processing principles that exploit their underlying physics.

\section*{Acknowledgement}

This work was partially undertaken when the authors belonged to the RIKEN Advanced Science Institute, which was reorganized and integrated into RIKEN as of the end of March, 2013. 
We thank Prof. Masahiko Hara for valuable discussions.

\end{document}